\documentclass[letterpaper, 10 pt, conference]{ieeeconf}  

\IEEEoverridecommandlockouts                              

\usepackage{graphics} 
\usepackage{epsfig} 
\usepackage{mathptmx} 
\usepackage{times} 
\usepackage{amsmath} 
\usepackage{amssymb}  
\usepackage{multirow}
\usepackage{booktabs}
\usepackage{hyperref}

\title{\LARGE \bf
Unveiling the Impact of Data and Model Scaling on High-Level Control for Humanoid Robots}

\author{
  Yuxi Wei$^{1,2}$  \quad Zirui Wang$^{2,3}$ \quad Kangning Yin$^{1,2}$ \quad
 Yue Hu$^{4}$ \\ Jingbo Wang$^{2*}$ \quad Siheng Chen$^{1*}$\\~
  \small $^1$ Shanghai Jiao Tong University \quad
  $^2$ Shanghai AI Laboratory \quad
  \small $^3$ Zhejiang University \quad $^4$ University of Michigan \\
  \texttt{\footnotesize wyx3590236732@sjtu.edu.cn}\\
  \href{https://vfishc.github.io/schur/}{\texttt{\small vfishc.github.io/schur}}
}

\begin{document}

\maketitle

\footnotetext[1]{Corresponding authors.}

\thispagestyle{empty}
\pagestyle{empty}

\begin{abstract}

Data scaling has long remained a critical bottleneck in robot learning. For humanoid robots, human videos and motion data are abundant and widely available, offering a free and large-scale data source. Besides, the semantics related to the motions enable modality alignment and high-level robot control learning. However, how to effectively mine raw video, extract robot-learnable representations, and leverage them for scalable learning remains an open problem. To address this, we introduce Humanoid-Union, a large-scale dataset generated through an autonomous pipeline, comprising over 260 hours of diverse, high-quality humanoid robot motion data with semantic annotations derived from human motion videos. The dataset can be further expanded via the same pipeline. Building on this data resource, we propose SCHUR, a scalable learning framework designed to explore the impact of large-scale data on high-level control in humanoid robots. Experimental results demonstrate that SCHUR achieves high robot motion generation quality and strong text-motion alignment under data and model scaling, with 37\% reconstruction improvement under MPJPE and 25\% alignment improvement under FID comparing with previous methods. Its effectiveness is further validated through deployment in real-world humanoid robot.
\end{abstract}

\section{INTRODUCTION}
Scaling up data and model size for scalable learning has led to significant breakthroughs across many research domains, including natural language processing, visual segmentation and detection, as well as image and video generation, and is now being extended to an increasing number of fields\cite{achiam2023gpt,zhang2022dino,kirillov2023segment}. In the context of robot learning, scaling has recently attracted substantial attention\cite{lin2024data}. However, for robots, directly collecting data is often complex, difficult, expensive, and time-consuming, which has motivated the search for broader, more automated, and low-cost data sources.

For humanoid robots, whose control and behaviors are highly similar to those of humans, human motion provides a valuable prior. Human motion datasets and related videos are comparatively easier to acquire at large scale, and their associated semantic information can be extracted through automated methods. This enables natural and efficient data expansion and scaling for humanoid robots by leveraging human motion as a foundation. Moreover, incorporating human motion priors facilitates more effective high-level control and cross-modal alignment for humanoid robots.
Meanwhile, semantics play a central role as an intermediate modality and as one of the most critical forms of control signals. Achieving robust semantic alignment under scaling is essential for complete high-level control and serves as a foundation for advancing toward diverse control paradigms, including vision-language-action models\cite{ding2025humanoid}.
\begin{figure}
    \centering
    \includegraphics[width=0.95\linewidth]{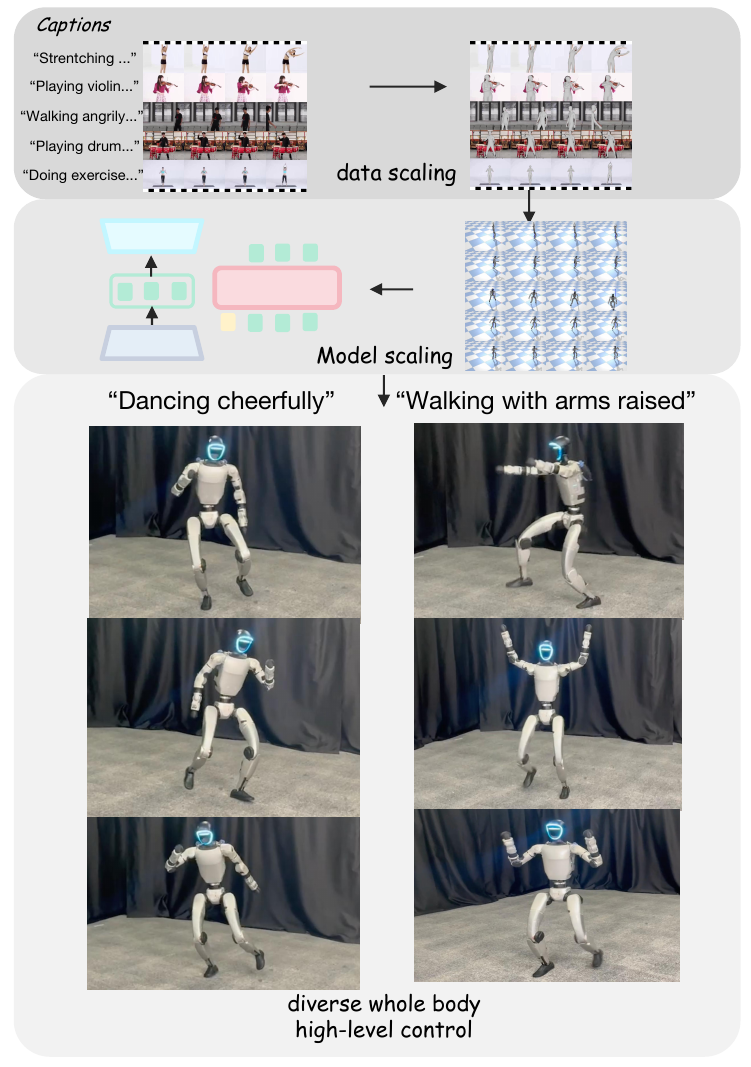}
    \vspace{-5mm}
    \caption{\small We present the large-scale, high-quality robot motion dataset \textbf{Humanoid-Union} to achieve data scaling. Additionally, we propose a text-based robot motion control method, \textbf{SCHUR}, which enables effective scalable learning. By scaling both data and models, our approach can generate complex, diverse and high-quality whole-body robot motions from text, which are executed by a whole-body tracker and successfully deployed on real-world robots.}
    \vspace{-4mm}
    \label{fig:teaser}
\end{figure}

Using human motion as prior for high-level control of humanoid robots has been explored recently~\cite{yue2025rl,shao2025langwbc,shi2025adversarial}. High-level control means generate control signal for low-level executor by conditions like texts. However, existing studies have shown several shortcomings. First, \cite{yue2025rl,ding2025humanoid} still rely on human motion as a direct prior, yet its distribution differs from robot motion. Although retargeting can transfer human to robot motion, it complicates usage and alters the distribution, thereby limiting modality alignment.
 Second, parts of human motion data may not be executable by low-level executors, reducing data quality and limiting execution effectiveness.
Third, existing works are limited in a small data scale. The commonly used dataset HumanML3D\cite{guo2022generating} contains only less than 30 hours of data. They fail to fully utilize the abundant data available currently, lacked corresponding model scaling, and have not fully tapped into the potential of human motion data for humanoid robots.


Therefore, it is essential to generate high-quality and executable robot motion not just human motion while fully leveraging the potential of large-scale human motion data. By expanding the data scale and correspondingly scaling the models, we can achieve more effective control capabilities and improved generalization.
Some studies have made preliminary explorations~\cite{mao2024learning}, but significant issues remain. First, it lacks investigation into robot motion representation, relying only on root information and DoFs without introducing more effective priors such as keypoints in Euclidean space for additional regularization. Second, it overlooks the advantages of directly generating robot motion and neglects to refine or filter robot motion data, leaving unexecutable or noisy segments unaddressed. Third, the exploration of scalable learning is shallow, with no effective methods or study of model scaling. Finally, they do not demonstrate lower-limb motions on real robots, instead focusing on arm movements, resulting in limited expressiveness.


To fully leverage large-scale human motion priors for humanoid robots and to investigate robot motion representations along with high-level scalable learning, it is crucial to construct a large and high-quality robot motion dataset automatically derived from human videos and motion, together with effective scalable learning strategy. To this end, we introduce the Humanoid-Union dataset and propose SCHUR, a scalable large text-based robot motion generation model.

The Humanoid-Union dataset comprises over 260 hours of source motion data, featuring an automatic pipeline and post-process for data scaling. Human-centric videos from various sources are converted into human motion through video motion capture techniques, while existing high-quality human motion data is also integrated and retargeted to obtain robot motion. Due to the extensive and varied sources of human motion data, which include interactions with scenes, variations in motion amplitude, and potential occlusions or noise, the robot motion is subsequently filtered and processed using a pretrained universal whole-body motion tracker. This ensures that only high-quality data, compliant with robotic kinematics and physical constraints, is retained at the robot motion level. Additionally, the descriptions corresponding to the motions are automatically annotated using a vision-language model, such as GPT-4V~\cite{achiam2023gpt}, enabling the construction of text-robot motion pairs for semantic alignment training and text generation to achieve high-level control.

To fully exploit the Humanoid-Union dataset, we propose SCHUR, an effective scalable learning model designed for high-quality text-based robot motion generation. SCHUR optimizes robot motion representation, and adopts an effective quantization method and related framework with proper training strategy. Although the root position, root orientation, and degrees of freedom (DoFs) can comprehensively represent the robot's state, we find that relying solely on these representations does not yield well generation results. Therefore, we further incorporate the positions and orientations of bound virtual keypoints calculated through forward kinematics (FK) into the representation. The redundant information provided by these virtual keypoints is similar to human motion style and effectively enhances the generation quality.
SCHUR adopts a two-stage approach: first, robot motion is tokenized, and then a LLaMA-based architecture is utilized for motion generation using text prefix. In the tokenization stage, we apply Finite Scalar Quantization (FSQ) VAE~\cite{mentzer2023finite}, which includes a more effective quantization method that significantly improves tokenization results compared to VQ-VAE and is less prone to collapse, thereby supporting larger codebook sizes. In the autoregressive generation stage, a prefix bidirectional attention mask is employed to generate the corresponding motion tokens, which are decoded to obtain the generated motion.

Through experiments, we validate the tokenization quality and scaling properties of SCHUR on the Humanoid-Union dataset. As the codebook size increases, there is a continuous improvement in the quality of tokenization, with FSQ demonstrating significantly better scaling capabilities compared to VQ. Additionally, in the second stage of the generator, an increase in model parameters also exhibited markedly improved modality alignment and generation quality. We further confirm the impact of high-quality data and advantage of directly generating robot motion on the success rate and tracking error of the low-level tracker execution, utilizing the tracker deployed on a real-world robot to verify the effectiveness in real-world robot scenarios.

\section{Related Works}
\subsection{Human Motion Generation}
Human motion generation has been extensively explored. Transformer-based~\cite{zhang2023generating, guo2024momask} and diffusion-based~\cite{tevet2022human, shafir2023human,xiao2025motionstreamer} methods have been proposed to achieve text-to-human-motion generation. Compared with robot motion data, human motion data are easier to obtain through various sources and benefit from numerous existing datasets~\cite{guo2022generating,mahmood2019amass}. However, directly using human motion data for high-level control of humanoid robots reduces modality alignment and complicates the control pipeline. Moreover, robot motion is subject to additional kinematic and physical constraints for deployment in the real world, making some human motion data unsuitable as priors for humanoid robots. In this work, training is performed directly on robot motion data, which are filtered and processed using a universal whole-body tracker. This procedure removes low-quality data while providing physical and kinematic priors, ensuring more effective and realistic learning for humanoid robot control.

\subsection{Humanoid Robot Learning}
Extensive research has focused on low-level control policies for humanoid robots. Prior work has explored full-body tracking, achieved loco-manipulation, and implemented text-based control via reinforcement learning (RL)~\cite{ji2024exbody2,he2024omnih2o,shao2025langwbc,yue2025rl,ze2025twist}. Nevertheless, RL-based low-level policies are difficult to scale with large datasets due to the constraints of lightweight policy model. Approaches that rely on human motion for robot require additional retargeting, which reduces modality alignment, complicates usage processes, and still fails to fully exploit data scalability. Methods using robot motion data often include infeasible motions and lack effective scalable learning strategies. In contrast, the Humanoid-Union dataset introduced in this work provides high-quality, tracker-processed robot motion data and SCHUR supports a more efficient approach to scalable learning.

\section{Humanoid-Union Dataset}
\begin{figure}
    \centering
    \includegraphics[width=0.95\linewidth]{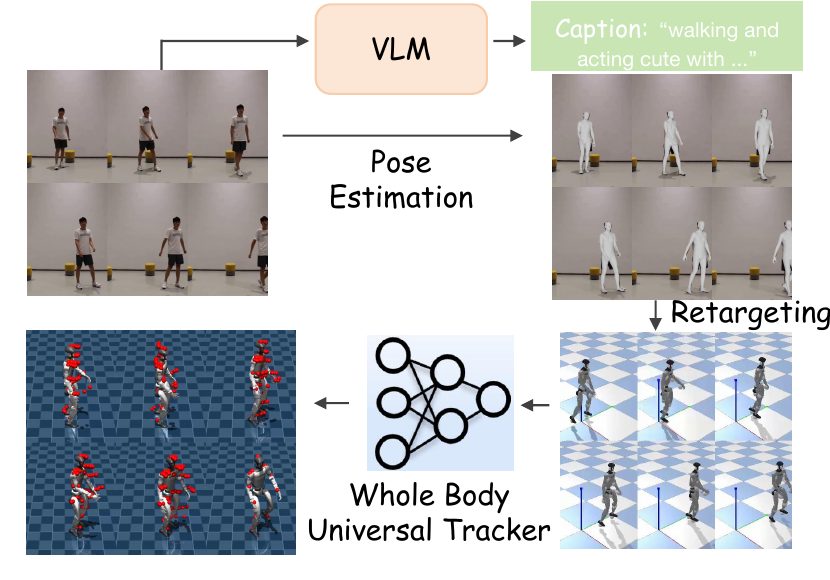}
    \vspace{-5mm}
    \caption{\small The data pipeline of \textbf{Humanoid-Union}. Pose data is extracted from videos by pose estimation, resulting in the SMPL representation. Corresponding descriptions are generated using vision-language model. The SMPL data is then retargeted to obtain the corresponding robot motion data, which is subsequently filtered and post-processed using a trained universal whole-body tracker.}
    \vspace{-3mm}
    \label{fig:data-process}
\end{figure}

\subsection{Overview}
Diverse and abundant human video and motion data are suitable sources for humanoid robots. However, directly applying human motion introduces large portions of data unsuitable for robots due to scene interactions, joint constraints, and other factors. Moreover, using human motion for high-level control complicates the process and limits modality alignment. To bridge this gap, we require a high-quality, large-scale robot motion dataset executable by robots. Furthermore, to enable high-level control, we provide textual descriptions for each motion, aligning robot motion with text as an essential and general intermediate modality.


We construct the Humanoid-Union dataset, which includes various sources~\cite{lin2023motion, liao2024animationgpt, lu2025scamo} and some self-collected data, resulting in about 260 hours of robot motion data, 170,000 sequences. We create the dataset with 3 steps: i)\textbf{video motion capture and description}, see sec~\ref{sec:mocap}. We extract human motion~\cite{lin2023motion} from a vast array of human-centric videos and employ a vision-language model to perform semantic labeling of the sequences, while also integrating existing publicly available human motion data with semantic annotations. ii) \textbf{retargeting}, see sec \ref{sec:retargeting}, human motion data is then retargeted to generate robot motion that conforms to kinematic constraints. iii) \textbf{whole body tracker filtering and postprocess}, see sec \ref{sec:tracker}, considering the wide-ranging sources of human-centric data, which may include a significant number of actions involving scenes and object interactions, as well as certain motions that may exceed the current capabilities of robots and the inherent noise in the data processing, we utilize a pretrained universal whole-body tracker for post-processing. Motions that fail during tracker execution are filtered out, and the motions processed by the tracker are used to complete the final post-processing. The data pipeline is shown in Fig.\ref{fig:data-process}
\begin{figure}
    \centering
    \includegraphics[width=0.85\linewidth]{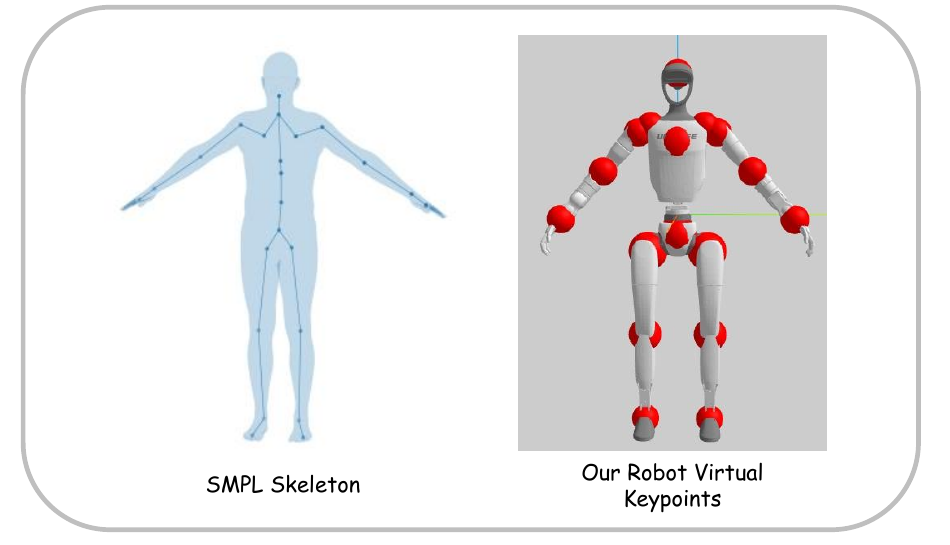}
    \vspace{-4mm}
    \caption{\small Comparison between the commonly used SMPL skeleton and our manually bound keypoints reveals that the virtual keypoints we defined exhibit a topology that is closer to SMPL. This alignment allows for a more accurate representation, maintaining a style that is more consistent with human motion.}
    \vspace{-3mm}
    \label{fig:keypoints}
\end{figure}
\subsection{Video MoCap and Description}
\label{sec:mocap}
Here, we extract human motion from videos and generates the corresponding textual descriptions with 3 steps. First, we detect and track human in the video sequences. Second, we perform pose estimation and refinement for each human motion sequence. Note that these two steps follow the standard procedure in ~\cite{lin2023motion}, and the output unified human motion data is represented in the SMPL model format~\cite{pavlakos2019expressive}, including the SMPL rotation parameters $\theta_{\text{human}}$ and the position $r^{\text{pos}}_{\text{human}}$. Third, we annotate the human motion data with additional textual descriptions via vision-language model~\cite{achiam2023gpt}. These three steps result in paired multi-modality data, enabling high-level control learning.

\subsection{Motion Retargeting}
\label{sec:retargeting}
Motion retargeting focuses on transforming human motion data into robot states. We bind some virtual keypoints to the robot by some fixed topology relationship that mimics the structure of the human skeleton, improving the alignment between human and robot topologies. The retargeting is achieved by mapping human motion data to the robot states including the virtual keypoints. The virtual keypoints also preserve properties such as symmetry, making the resulting motions more human-like in style and also benefits the following model learning as part of representation. Note that, these points are not in a one-to-one correspondence with the robot’s physical parts but instead satisfy specific relative relationships. The virtual keypoints is shown in Fig.\ref{fig:keypoints}
 
Specifically, motion retargeting is achieved with 4 steps: i) bind virtual keypoints for the robot by some artificial definition of relationship with body parts; ii) recalibrate the scale of the body parts using a fixed T-pose; iii) based on the inverse kinematics (IK) process, we solve for the robot’s joint positions $j_{\text{robot}}$; iiii) and then employ forward kinematics (FK) to compute the virtual keypoints’ positions $k^{\text{pos}}$ and rotations $k^{\text{ori}}$. The height of the robot motion is corrected using a strategy similar to PHC~\cite{luo2023perpetual}, followed by a smoothing post-processing step.

The final results include the robot's root position $r^{\text{pos}}_{\text{robot}} \in \mathbb{R}^{3}$, root orientation $r^{\text{ori}}_{\text{robot}} \in \mathbb{R}^{3}$, joint positions for all DoFs $j_{\text{robot}} \in \mathbb{R}^{d}$, and the positions $k^{\text{pos}} \in \mathbb{R}^{n \times 3}$ and rotations $k^{\text{ori}} \in \mathbb{R}^{n \times 3}$ of virtual keypoints. Here, $d = 29$ is the dimensionality of the DoFs, and $n = 17$ is the number of virtual keypoints. $r^{\text{ori}}_{\text{robot}}$ and $k^{\text{ori}}$ are represented with roll-pitch-yaw.


\subsection{Whole Body Tracker and Postprocessing}
\label{sec:tracker}
We train a whole-body tracker for both advanced data processing and final low-level motion execution. The tracker is trained on the almost full AMASS dataset~\cite{mahmood2019amass}. After extensive training, the resulting universal tracker is capable of directly imitating whole-body motions, including lower-limb movements. The training pipeline resembles that of ExBody2~\cite{ji2024exbody2}, where the retargeted motion results, including the virtual keypoints, are used as observations.
This tracker demonstrates strong generalization and a high success rate in execution. It is employed to further filter and refine robot motion data by leveraging the execution constraints, discarding sequences that are evidently infeasible while preserving motions better suited for robotic execution. About 10\% of motions are filtered. In addition, the tracker introduces physics-informed priors into the motion representations. Finally, we deploy the same tracker in our real-world robot experiments, validating its effectiveness.

\section{SCHUR for Scalable Robot Motion Generation}
\begin{figure}
    \centering
    \includegraphics[width=0.99\linewidth]{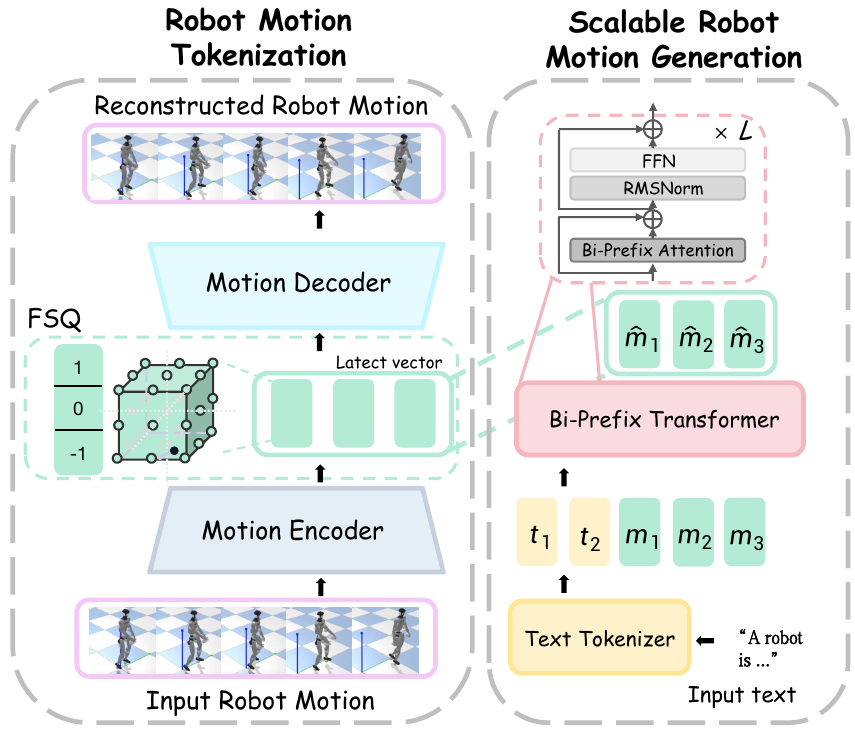}
    \vspace{-5mm}
    \caption{\small The framework of \textbf{SCHUR} consists of two stages. In the first stage, robot motion is tokenized using an effective representation, with FSQ employed for quantization. Here, we present an example using L=3 to illustrate FSQ. The second stage utilizes text tokens as prefix and applies prefix-bidirectional attention to generate motion tokens in an autoregressive manner.}
    \label{fig:method}
    \vspace{-3mm}
\end{figure}
\subsection{Overview}
With a large-scale, high-quality robot motion dataset, we focus on effectively leveraging it by aligning robot motion with semantic descriptions. Text, as a key intermediate modality, enables high-level control, and this alignment underpins the Vision-Language-Action (VLA) model. Building on the Humanoid-Union dataset, we propose \textbf{SCHUR} (\textbf{Sc}alable \textbf{Hu}manoid \textbf{R}obot model), which scales to achieve text-based high-level control of humanoid robots. SCHUR draws on successful experiences from computer vision~\cite{razavi2019generating}, employing a paradigm of discrete tokenization and autoregressive generation. It consists of two stages. In the first stage, the representation of the motion is tokenized to obtain a compact and structured discrete representation. We improve the representation of robot motion itself, effectively enhancing the quality of the tokenizer, and apply FSQ, a more effective quantization method, which further improves quality while enhancing the model's scaling capability, reducing the likelihood of collapse. In the second stage, we adopt a decoder-only autoregressive generation approach based on a LLaMA~\cite{touvron2023llama} structure, using the tokens from text as prefixes to autoregressively generate the robot motion tokens from the first stage. The overview of SCHUR is in Fig.\ref{fig:method}

\subsection{Robot Motion Representation}
Robot motion itself can be accurately and completely represented through the root position $r^{pos}_{robot}$, root orientation $r^{ori}_{robot}$, and degrees of freedom (DoFs) $j_{robot}$, which is the approach taken in \cite{mao2024learning}. However, we find that such a representation does not yield sufficiently effective tokenization or satisfactory generation results, the related results can be found in Tab.~\ref{tab:performance_comparison}. To address this, we introduce a more enriched representation. During the retargeting process, the virtual keypoints we bind can effectively restore the human motion style and additionally provide a representation of the motion within the 3D Euclidean space. We combine the positions $k^{pos}$ and orientations $k^{ori}$ of the virtual keypoints into the representation, resulting in a final representation $e = \text{concat}(r^{pos}_{robot}, r^{ori}_{robot}, j_{robot}, k^{pos}, k^{ori})$. Experimental results demonstrate that this redundant representation of robot motion is highly effective for tokenization.


\subsection{Scalable and Efficient Motion Tokenizer}
The preliminary representation of robot motion must be tokenized into discrete tokens for subsequent generation stage. Effective tokenization also facilitates broader applications in modality alignment and high-level control, such as within VLA frameworks. Previous approaches typically adopt VQ-VAE for this stage, and discrete tokenization has also achieved success in many generative tasks. However, as the scale of data increases, VQ-VAE requires a correspondingly larger codebook to capture the expanded distribution. Due to the quantization mechanism of conventional VQ, particularly the backpropagation process, such methods are prone to model collapse, making it difficult to achieve sustainable scaling.
To overcome these limitations, we adopt Finite Scalar Quantization (FSQ), a more robust quantization strategy that mitigates collapse while enabling more effective tokenization and continued scaling.

\textbf{Conventional VQ-VAE.} VQ-VAE can be employed to learn a discretized representation of robot motion, following the standard encoder–decoder paradigm. The input motion $e$ is first encoded into a latent vector $z$, which is then quantized by the operator $\textit{Q}$ to obtain a discrete latent vector $\hat{z}$. Finally, $\hat{z}$ is decoded back into the reconstructed robot motion $\hat{e}$.  

In the conventional VQ-VAE formulation, a codebook $\textit{C} = \{c_{i}\}_{i=1}^{S}$ with size $S$ is learned for quantization. The quantization process can be expressed as:
\[
\hat{z} = \textit{Q}(z; \textit{C}) = \mathop{\arg\min}\limits_{c_{k}} \| z - c_{k} \|_{2}^{2}.
\]
The optimization simultaneously considers the reconstruction quality of the decoder and the learning of the codebook. The overall loss function is written as:
\[
\mathcal{L} = \| e - \text{Dec}(\textit{Q}(z; \textit{C})) \|_{2}^{2} 
+ \alpha \| z - \text{sg}(\hat{z}) \|_{2}^{2},
\]
where $\text{Dec}$ denotes the decoder, $\alpha$ is a weight parameter, and $\text{sg}(\cdot)$ indicates the stop-gradient operation.  

To stabilize training, techniques such as exponential moving average (EMA) and codebook reset, originally proposed for VQ-VAE in human motion modeling~\cite{zhang2023generating}, are also adopted in our framework. However, in order to handle larger-scale datasets and achieve higher compression quality, it is necessary to increase the codebook size. A common issue with conventional VQ-VAE is codebook collapse when the codebook becomes large, since the $\arg\min$ operation in VQ often tends to select only a subset of codes for frequent updates, leaving other codes underutilized. To address this, we adopt a more effective quantization strategy, named FSQ.


\begin{table}[tbp]
\centering
\caption{Ablation of tokenizer}
\vspace{-3mm}
\label{tab:performance_comparison}
\begin{tabular}{c|cccc}
\toprule
Metric & \multicolumn{1}{c}{Human Motion} & \multicolumn{1}{c}{20\% Data} & \multicolumn{1}{c}{Naive Repre} & \multicolumn{1}{c}{Ours} \\
\midrule
MPJPE & 0.0389 & 0.0483 & 0.0406 & \textbf{0.0326} \\
MPKPE & 0.0401 & 0.0497 & 0.0454 & \textbf{0.0372} \\
L1 Loss & 0.0413 & 0.0512 & 0.0463 & \textbf{0.0380} \\
\bottomrule
\end{tabular}
\vspace{-2mm}
\end{table}

\textbf{Finite Scalar Quantization(FSQ).} To address the potential collapse caused by the argmin operation, we adopt FSQ, a more effective quantization method that avoids using argmin. In FSQ, the quantization process can be written as:
\[
\hat{z} = \textit{Q}(z) = \text{round}(f(z)),
\]
where $f(\cdot)$ is a bounding function, which in our implementation is set as the sigmoid function. Unlike VQ, FSQ does not involve the $\arg\min$ operation, thus alleviating the collapse problem inherent in VQ.  

FSQ essentially quantizes each element of the latent vector into one of $L$ unique integers, and the number of different combinations and the codebook size can be expressed as:
$|\textit{C}| = \prod_{i=1}^{d} L_{i},$
where $d$ is the dimensionality of the latent vector. Since the rounding operation in the quantization process is non-differentiable, we also employ a similar stop-gradient technique.  

The optimization in FSQ does not require additional regularization or optimization strategies; it directly optimizes the reconstruction error from the decoder. Therefore, the final loss function is written as:
\[
\mathcal{L} = \| e - \text{Dec}(f(z) + \text{sg}(\text{round}(f(z)) - f(z))) \|_2^2.
\]
FSQ does not require modifications to the encoder or decoder, only changes to the quantization process itself. As a result, FSQ achieves significantly better reconstruction quality and scaling properties.


\subsection{Autoregressive Text-Based Generation}
Using the high-quality robot motion tokens obtained through FSQ, we aim to align with the text modality, facilitating the generation of robot motion from text. The overall generation process adopts a decoder-only transformer architecture similar to LLaMA. The robot motion tokens are already acquired from the previous stage, while the text tokens are obtained from a pre-trained T5-XL large language model~\cite{raffel2020exploring}, which provides word-level tokens as the sequence prefix. These text tokens are then used for autoregressive generation of robot motion tokens.  

It is worth noting that since text tokens do not require prediction, we modify the attention mask to enable prefix-bidirectional attention, in contrast to the conventional causal attention. This approach allows bidirectional attention over the text tokens while performing autoregressive prediction over the motion tokens. Additionally, RMSNorm is applied within the transformer blocks to improve training stability.  

The final optimized loss can be written as:
\[
\mathcal{L} = -\sum_{t=1}^{n} \log p(\hat{m_t} | m_{< t}, T),
\]
where $T$ represents the text token series and $m$ denotes the motion token.


\begin{table}[tbp]
\centering
\caption{Comparison and ablation of text-based generation}
\vspace{-3mm}
\label{tab:performance_comparison_2}
\begin{tabular}{c|cccc}
\toprule
Metric & \multicolumn{1}{c}{Human Motion} & \multicolumn{1}{c}{20\% Data} & \multicolumn{1}{c}{MDM} & \multicolumn{1}{c}{Ours} \\
\midrule
FID & 16.9 & 23.5 & 20.8 & \textbf{12.6} \\
R@1 & 0.688 & 0.653 & 0.667 & \textbf{0.719} \\
R@2 & 0.717 & 0.689 & 0.692 & \textbf{0.754} \\
R@3 & 0.755 & 0.707 & 0.713 & \textbf{0.812} \\
\bottomrule
\end{tabular}
\vspace{-1mm}
\end{table}

\begin{table}[tbp]
\centering
\caption{tracking performance under different high-level control}
\vspace{-3mm}
\label{tab:performance_comparison_3}
\begin{tabular}{c|c|ccc}
\toprule
Metric & \multicolumn{1}{c}{Dataset} & \multicolumn{1}{c}{Human Motion} & \multicolumn{1}{c}{W. raw data} & \multicolumn{1}{c}{Ours} \\
\midrule
Success Rate & 0.911 & 0.774 & 0.762 & \textbf{0.907} \\
MPJPE & 0.0874 & 0.1183 & 0.1095 & \textbf{0.0893} \\
MPKPE & 0.0867 & 0.1112 & 0.1028 & \textbf{0.0878} \\
\bottomrule
\end{tabular}
\vspace{-3mm}
\end{table}

\section{Experiment}

\begin{figure*}
    \centering
    \includegraphics[width=0.95\linewidth]{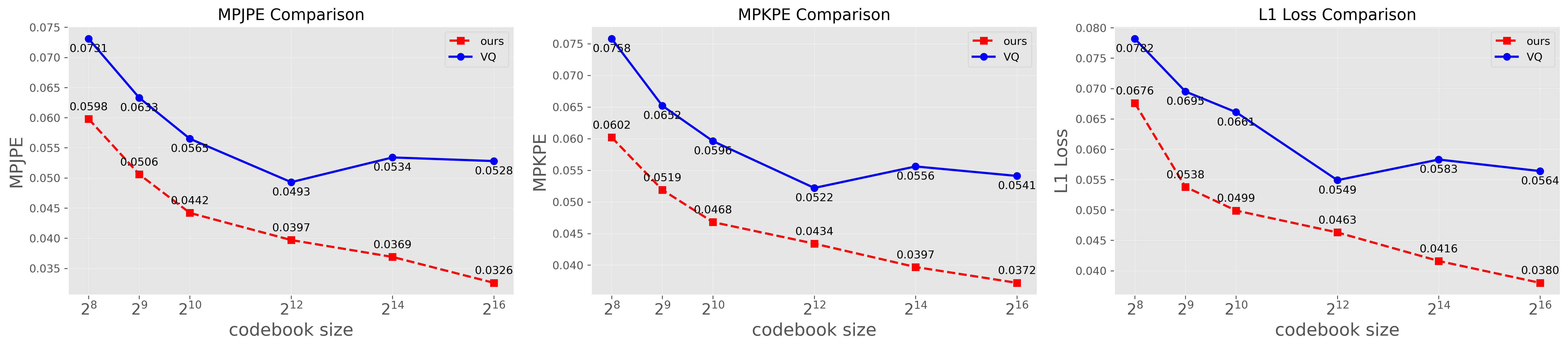}
    \vspace{-3mm}
    \caption{\small Comparison of SCHUR's tokenization stage with conventional VQ-VAE under different codebook sizes, using metrics such as MPJPE, MPKPE, and L1 loss. With the introduction of FSQ, SCHUR significantly outperforms conventional VQ-VAE in terms of reconstruction quality. Furthermore, it demonstrates strong scaling properties, with performance continuously improving as the codebook size increases. In contrast, VQ-VAE experiences instability during training with larger codebook sizes and eventually suffers from collapse.}
    \vspace{-2mm}
    \label{fig:stage1}
\end{figure*}

\begin{figure*}
    \centering
    \includegraphics[width=0.99\linewidth]{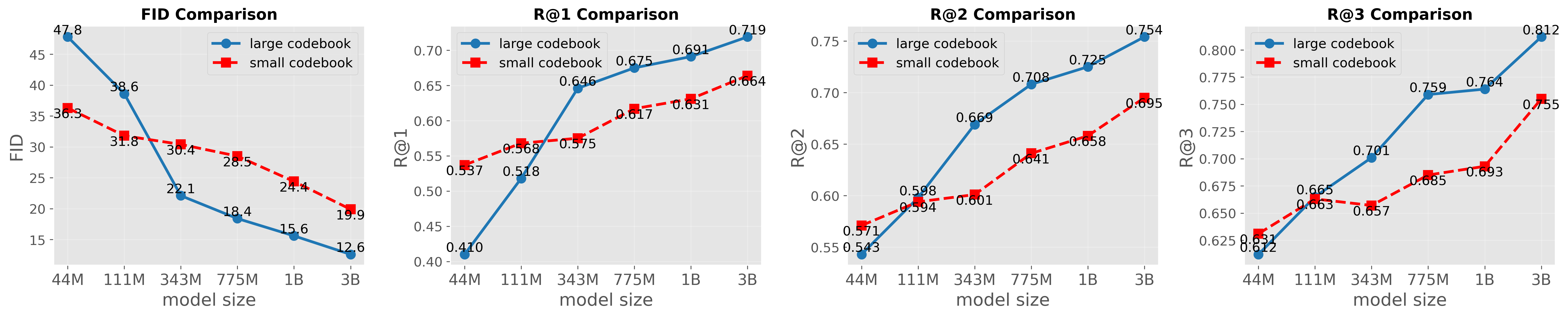}
    \vspace{-3mm}
    \caption{\small In the text-based generation stage, the FID, R@1, R@2, and R@3 metrics are compared under small (256) and large (65536) codebook sizes, with the model's parameter size increasing. As the model size grows, SCHUR is able to better leverage the data, improving its alignment with the text modality across all metrics. Additionally, the larger codebook size, due to its superior tokenization performance, generally leads to better generation results. However, when the model parameters are small, the larger codebook size struggles to learn effectively, resulting in poorer performance.
}
    \vspace{-3mm}
    \label{fig:stage2}
\end{figure*}

\subsection{Experiment setup}
All our experiments are conducted on the 29-DoF version of the Unitree G1. In the first stage, the tokenizer employs convolutional residual blocks for both the encoder and the decoder. In the second stage, for text-based autoregressive generation, we adopt an architecture that is nearly identical to LLaMA, with model parameter scales aligned accordingly. 
Our tracker is fully trained on the AMASS dataset with IssacGym~\cite{makoviychuk2021isaac} as simulator, following data process similar to PHC~\cite{luo2023perpetual}. The tracker is universal in design and does not undergo additional fine-tuning for specific motion. MuJoCo~\cite{todorov2012mujoco} is used for evaluation. Out of the 29 degrees of freedom, 6 are fixed, leaving 23 actively controlled in practice. Detailed configurations for each component are presented in their respective sections.

\begin{figure*}
    \centering
    \includegraphics[width=0.95\linewidth]{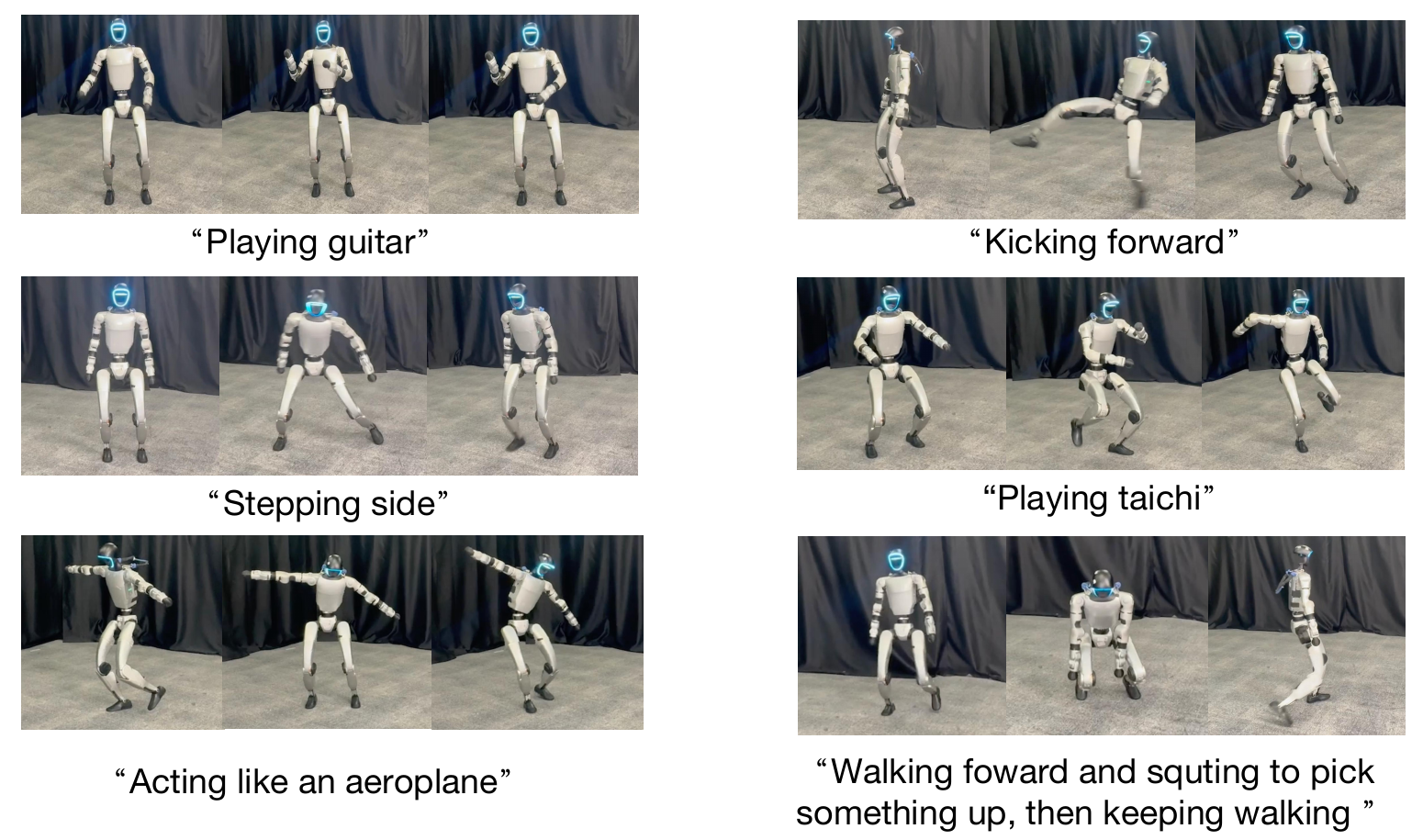}
    \vspace{-3mm}
    \caption{\small The real-world deployment results across diverse textual inputs. SCHUR is capable of handling a wide range of requirements, including upper- and lower-limb movements, locomotion, abstract semantic control, and complex long-horizon control. It generates the corresponding whole-body information as control signals, which are then executed with high precision and stability through a universal whole body tracker, thereby achieving smooth and reliable real-world deployment performance. See more results in supplementary video.}
    \label{fig:real-world-robot}
    \vspace{-3mm}
\end{figure*}

\subsection{Model scaling of tokenizer}
We conduct scaling experiments on the tokenizer under varying codebook sizes and compared its performance against a conventional VQ-VAE baseline, as shown in Fig.\ref{fig:stage1}. The evaluation metrics include MPJPE (Mean Per Joint Position Error), MPKPE (Mean
Per Keypoints Position Error), and L1 loss. In our experiments, the Humanoid-Union dataset was partitioned into training, test, and validation sets with an 80\%, 15\%, and 5\% split, respectively, and results are reported on the test set.
From the results, we observe that our method, when incorporating FSQ, consistently outperforms the conventional VQ approach across all reconstruction metrics. This provides strong evidence that our tokenizer is more effective, achieving superior performance even under relatively small codebook sizes.

In addition, we observe that as the codebook size increases, conventional VQ exhibits an obvious collapse effect, failing to further improve reconstruction performance. In contrast, our method continues to achieve stable gains in reconstruction quality with larger codebook sizes, demonstrating its superior scalability. To further validate the hypothesis proposed in our paper, we evaluated the codebook usage of different methods, defined as the proportion of the codebook utilized on the test set. Our method consistently achieves a usage rate above 99\% across all settings, whereas conventional VQ shows a gradual decline to below 90\% as the codebook size grows. Overall, the tokenization in SCHUR is shown to be highly effective and to possess strong scaling properties.

\subsection{Ablation of tokenizer}
We further conducted an ablation study on the tokenizer under the largest codebook size, comparing several settings: (i) directly training on human motion and subsequently applying retargeting to robot motion for evaluation, (ii) training with 20\% of full dataset and (iii) training with representations that only include root and DoF information, without incorporating virtual keypoints, where metric computation is performed after the FK process. As shown in Tab.\ref{tab:performance_comparison} direct training on human motion fails to achieve optimal performance due to distributional discrepancies and also complicates the practical deployment pipeline. Moreover, when the training data is limited, the performance degrades significantly, highlighting the importance of data scaling in the tokenization process. Finally, the inclusion of virtual keypoints in the representation substantially improves the reconstruction quality of the tokenizer.

\subsection{Model scaling of text-based generation}
We further evaluate the scaling characteristic of the generation stage under different codebook sizes (256 and 65,536) in the first stage, as shown in Fig.\ref{fig:stage2}. The dataset partitioning strategy is kept consistent with that used for the tokenizer. For evaluation, we employ four common metrics for generation: FID, R@1, R@2, and R@3. Following a procedure from \cite{xiao2025motionstreamer}, we train an evaluator on our own dataset.
The results indicate that as the model size increases, all metrics consistently improve, suggesting that larger models achieve better alignment between text and generated motion across all criteria, thereby validating the scaling capability of the generation stage. However, with smaller codebook sizes in the first stage, the relatively weaker reconstruction quality limits the ability to fully capture the data distribution, resulting in overall poorer generation quality. Interestingly, when the second-stage model size is very small, smaller codebook sizes sometimes yield relatively better results, as the limited model capacity prevents effective learning from large-scale data.
Overall, these results demonstrate that SCHUR is able to effectively exploit large-scale data through scaling, achieving high-quality generative performance.

\subsection{Comparison and ablation of text-based generation}
We conduct comparisons under the largest model configuration between three settings: (i) generating human motion and retargeting it to robot motion, (ii) training with only 20\% of the full data, and (iii) generation using MDM~\cite{tevet2022human}, a diffusion-based approach. As shown in Tab.\ref{tab:performance_comparison_2}, consistent with tokenizer experiments, directly generating human motion and subsequently applying retargeting is constrained by distributional distribution, which limit performance. Likewise, insufficient training data prevents the model from fully exploiting its capacity. Moreover, compared with MDM, SCHUR achieves superior generative performance, further demonstrating the effectiveness of our approach.

\subsection{Tracker performance under different high-level control}
To further verify the generation quality, we evaluate whether the universal tracker can reliably execute the generated motions, as shown in Tab.\ref{tab:performance_comparison_3}. 
We adopt success rate, MPJPE, and MPKPE as evaluation metrics. As a reference, we first validate the tracker's performance on full dataset. We then compare two alternative settings: (i) generating human motion followed by retargeting, and (ii) training SCHUR directly on raw robot motion data without our tracking-based filtering and post-processing.

The results show that in both alternative cases, the tracker’s success rate drops significantly. This is due to the presence of a large amount of infeasible or physically inconsistent motions in the training data, which contaminate the learning process. The increased execution difficulty also leads to degraded tracking performance. By contrast, our approach leverages higher-quality data, where unreasonable or overly difficult motions have been filtered out, resulting in a distribution more compatible with the tracker’s capabilities.

\subsection{Real-world deployment}
We further demonstrate the generated results on a real-world robot in Fig.\ref{fig:real-world-robot}, with additional results provided in the supplementary video. In combination with our tracker for deployment, the generative model is capable of producing motions that can be reliably executed on the real-world robot. These include, but are not limited to, directly controlled upper- and lower-limb actions, locomotion behaviors, as well as non-directly controlled motions with clearly abstract semantics. Moreover, SCHUR can compose multiple atomic actions into longer sequences, enabling versatile behavior generation. Our final deployment achieves highly precise and stable control, maintaining balance and motion stability even under whole-body tracking, while successfully executing a wide range of locomotion tasks.

\subsection{Conclusion}
We propose Humanoid-Union, a large-scale and diverse robot motion dataset with semantic information for modality alignment and control. This dataset provides high-quality robot motion, filtered through a universal whole-body tracker to remove unsuitable motions. To leverage this data, we introduce SCHUR, a scalable and effective method for high-quality, text-based robot motion generation. SCHUR combines FSQ-based tokenization and autoregressive generation with a text prefix, and adopts improved robot motion representation. Experiments demonstrate SCHUR’s superior performance, robust scaling capabilities, and the value of large-scale data and direct robot motion usage. Real-world deployment on real-world robot further validates its generation and control capabilities. As a limitation, we explore the direct control by robot states and actions based on text without low-level tracker. But large models face challenges in real-time deployment, and the lack of balance priors makes locomotion control difficult, which is left as future work .

\bibliographystyle{IEEEtranS}
\bibliography{root}

\end{document}